\documentclass{article}
    \PassOptionsToPackage{numbers, compress}{natbib}

    \usepackage[final]{neurips_2023}

\usepackage[utf8]{inputenc} 
\usepackage[T1]{fontenc}    
\usepackage{hyperref}       
\usepackage{url}            
\usepackage{booktabs}      
\usepackage{amsfonts}      
\usepackage{nicefrac}       
\usepackage{microtype}     
\usepackage{graphicx}
\usepackage{multirow}
\usepackage{graphicx}
\usepackage{xcolor,colortbl}
\usepackage{placeins}
\usepackage{float}
\usepackage{caption}
\usepackage{subcaption}
\usepackage{indentfirst}
\usepackage{amsmath}
\usepackage{wrapfig}
\usepackage{cleveref}
\usepackage{hyperref}
\usepackage{svg}
\usepackage{booktabs}
\usepackage{lipsum} 
\usepackage{blindtext}
\usepackage{etoolbox}

\makeatletter %
\patchcmd{\maketitle}%
  {\def\@makefnmark{\rlap{\@textsuperscript{\normalfont\@thefnmark}}}}%
  {\def\@makefnmark{\rlap{\@textsuperscript{\normalfont\color{red}\@thefnmark}}}}%
  {}
  {}
\makeatother

\title{Diffusion-C: Unveiling the Generative Challenges of Diffusion Models through Corrupted Data}

\author{%
Keywoong Bae$^{1,2}$\thanks{Work started while being student at Inha University} \quad Suan Lee$^{3}$ \quad Wookey Lee$^{1}$\thanks{Corresponding author} \\
$^1$Inha University \quad $^2$POSTECH \quad $^3$Semyung University\\
\texttt{kwbae@postech.ac.kr; }
\texttt{suanlee@semyung.ac.kr; }
\texttt{trinity@inha.edu;}
}

\begin{document}

\maketitle

\begin{abstract}
In our contemporary academic inquiry, we present "Diffusion-C," a foundational methodology to analyze the generative restrictions of Diffusion Models, particularly those akin to GANs, DDPM, and DDIM. By employing input visual data that has been subjected to a myriad of corruption modalities and intensities, we elucidate the performance characteristics of those Diffusion Models. The noise component takes center stage in our analysis, hypothesized to be a pivotal element influencing the mechanics of deep learning systems. In our rigorous expedition utilizing Diffusion-C, we have discerned the following critical observations: \textbf{(\uppercase\expandafter{\romannumeral1})} Within the milieu of generative models under the Diffusion taxonomy, DDPM emerges as a paragon, consistently exhibiting superior performance metrics. \textbf{(\uppercase\expandafter{\romannumeral2})} Within the vast spectrum of corruption frameworks, the fog and fractal corruptions notably undermine the functional robustness of both DDPM and DDIM. \textbf{(\uppercase\expandafter{\romannumeral3})} The vulnerability of Diffusion Models to these particular corruptions is significantly influenced by topological and statistical similarities, particularly concerning the alignment between mean and variance. This scholarly work highlights Diffusion-C's core understandings regarding the impacts of various corruptions, setting the stage for future research endeavors in the realm of generative models.

\end{abstract}

\section{Introduction}
Deep learning models have shown remarkable performances in various domains but unlike human vision, they are fundamentally vulnerable to adversarial attacks and various corruptions\cite{goodfellow2014explaining,silva2020opportunities,liu2019vulnerability}. 
Deep learning is significantly affected not only by substantial and critical noise but also by minor deformations.
Numerous benchmarks exist to identify the different failure patterns of models when subjected to a range of corruptions\cite{hendrycks2019benchmarking,mu2019mnist}.  
The Denoising Diffusion Probabilistic Model (DDPM)\cite{ho2020denoising,cao2022survey,croitoru2023diffusion}, which falls under the diffusion generative model category, has proven its prowess in creating top-tier images. 
Of late, DDPM has garnered significant attention, leading to numerous studies exploring its inherent boundaries.
Although many studies have been conducted \cite{DBLP:journals/access/TiagoSSM23, DBLP:conf/cvpr/ZhuangZL23, DBLP:conf/iclr/KimOY23, DBLP:conf/iclr/WuWPNX23}, only a handful have investigated the resilience of Diffusion Models in the context of vulnerabilitiy challenges.

Drawing from prior studies on the susceptibilities of Diffusion Models\cite{zhang2022gddim,kim2022diffusionclip, lu2022dpm,wang2023learning,li2022efficient,wallace2023edict,matsumoto2023membership}, this manuscript utilizes adversarial attack methodologies to probe the vulnerabilities inherent in Diffusion Models.
By training Diffusion Models using various corruptions both in type and intensity as input, our objective is to determine the specific disturbances to which the generative models are susceptible and their respective magnitudes.
For the purpose of this research, we've termed this approach of using adversarial attacks on Diffusion Models to evaluate how changes in input values influence output value shifts as \textbf{Diffusion-C}.

\section{Methodology}
\subsection{What is Diffusion-C?}

Original Diffusion Model's training mechanism consists of two processes, Diffusion and Denoising. 
In diffusion process, the original image ($x_0$) gradually transforms into a purely noisy sample ($x_T$) by being injected gaussian noises $(\beta_t)$ incrementally.
After then, DDPM returns the final image $x_g$ as a result of denoising process, 
while DDIM samples from intermediate noisy versions $q(\tilde{x}_t |x_0, x_k)$ of the data and then denoises from there.
If the state of $x_g$ and $x_0$ is similar, we interpret it as a good return performance of the Diffusion Models on their generative ability.

\begin{figure}[h!]
  \centering
  \includegraphics[width=\textwidth]{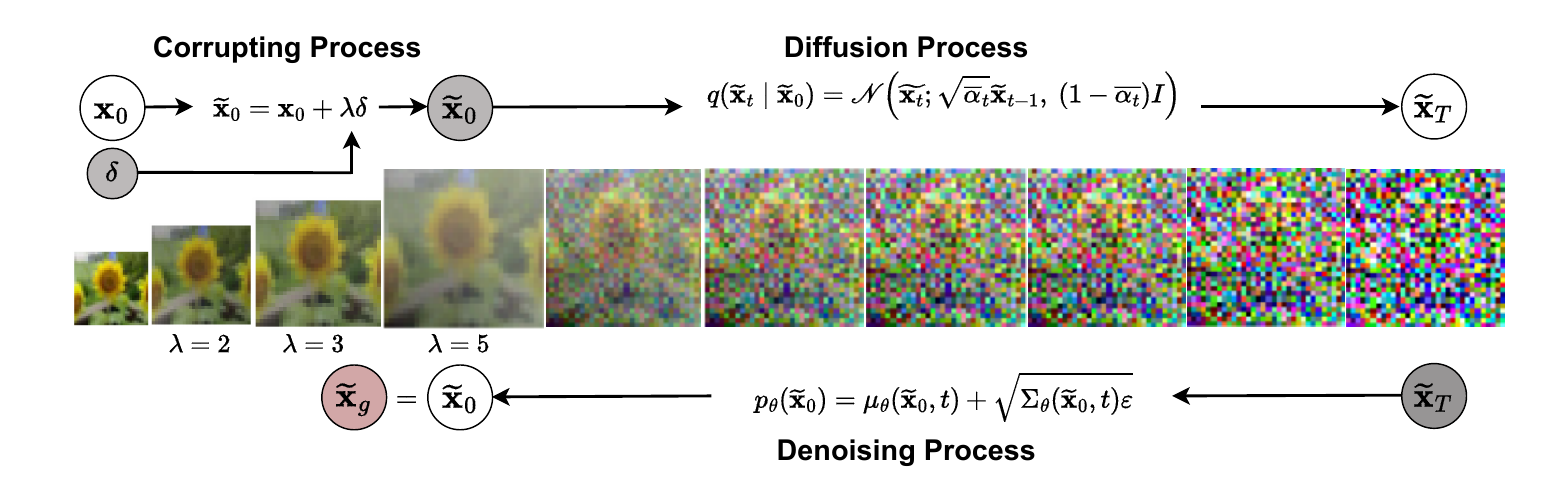}
  \caption{\textbf{Brief mechanism on Diffusion-C.} Unlike the mechanism of original Diffusion Models, Diffusion-C has a process of Corrupting. To summarize the generative limitations of Diffusion Models, we used corrupted images as an input and they were corrupted with various types and severities of corruption at corrupting process vice versa.}
  \label{fig:figure1}
\end{figure}

To observe the generative draws of Diffusion Models, we implemented a method of \textbf{Diffusion-C} through the following two stages of questions.

\textbf{$\mathbf{Q1}$) If the original initial input is corrupted with non-Gaussian noise $(\delta)$ instead of Gaussian noise $(\epsilon)$, can Diffusion Models still return quality images?}

At first, we contemplated whether the return performance of Diffusion Models would still be good in the case of substituting original input $(x_0)$ to corrupted input $(x_t)$ with some Gaussian noise $(x_0 + \epsilon=x_t)$.
And this inquiry was expanded based on the recent researches on corruption benchmarks\cite{hendrycks2019benchmarking,mu2019mnist} and various efforts for replacing  Gaussian noise to non-Gaussian noise\cite{nachmani2021non, daras2022soft}.
We replaced Gaussian noise to non-Gaussian noise, which is defined as Corruptions in this work.

We aimed to observe the degradation of Diffusion Models's return performance when $\tilde{x}_0 (=x_t + \delta)$ is used as an input instead of original $x_0$.
In \ref{comparison_exp}, we compared the return performance of Diffusion Models with those of DCGAN\cite{radford2015unsupervised} and WGAN-GP\cite{gulrajani2017improved}.

\textbf{$\mathbf{Q2}$) How far can Diffusion Models's Return performance degrade?}

This study primarily seeks to understand the boundaries of generative capabilities in Diffusion Models. By intensifying the corruption levels, we assessed the extent to which the performance of these models deteriorates.
After finding a specific corruption $\delta$, to which Diffusion Models are vulnerable, we observed how far Diffusion Models's performance will degrade with respect to the severity ($\lambda$) of the corruption.

\section{Experimental Results}
The image datasets, originally with values between 0 and 255, were resized to fall between -1 and 1 for training. We utilized the Adam optimizer, set with a learning rate of $10^{-4}$ and a batch size of 128. We incrementally adjusted the Gaussian noise by 0.02, beginning at $10^{-4}$. The model's performance was gauged using the Fréchet Inception Distance (FID) \cite{heusel2017gans}. This investigation sequentially built on the findings of each experiment, amplifying their significance, before advancing to the subsequent study. 

\subsection{Basic Comparison Experiments}\label{comparison_exp}
In our research, our initial experiments were focused on (i) assessing the return performance of the generative model, and (ii) pinpointing its most critical vulnerabilities to corruptions. We made comparative evaluations using models like DDPM\cite{ho2020denoising}, DDIM\cite{song2020denoising}, DCGAN\cite{radford2015unsupervised}, and WGAN-GP\cite{gulrajani2017improved}. From the results, DDPM stood out as the top performer, followed by DDIM, WGAN-GP and DCGAN. On average, DDIM trained 730/56 times faster than DDPM, but the quality of the images it produced was slightly lower than DDPM. Moreover, of all the corruption types examined, Diffusion Models were particularly vulnerable to fog corruption.

\begin{figure}[h]
  \centering
  \includegraphics[width=\textwidth]{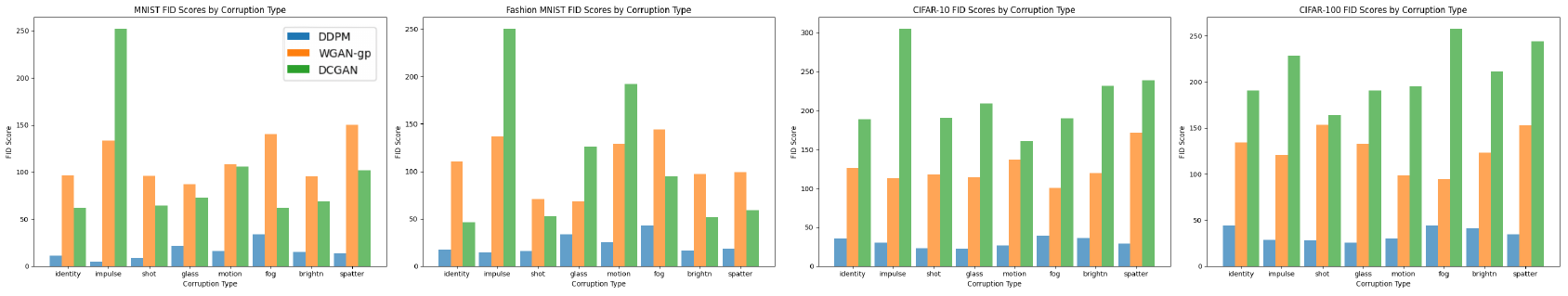}
  \caption{\textbf{Result of Fréchet Inception Distance score.} Blue is for DDPM, Orange for WGAN-GP, and Green for DCGAN.}
  \label{fig:figure6}
\end{figure}

\begin{table}[h!]
\centering
\caption{FID scores of DDPM/DDIM on the various corruptions}
\resizebox{\columnwidth}{!}{%
\begin{tabular}{@{}c|ccccccccc@{}}
\toprule
                      &               & Clear    & \multicolumn{2}{c}{Noise} & \multicolumn{2}{c}{Blur} & \multicolumn{2}{c}{Weather} & Extra   \\
                      &               & Identity & Impulse      & Shot       & Glass      & Motion      & Brightness      & \cellcolor[HTML]{BFBFBF}Fog       & Spatter \\ \midrule
\multirow{4}{*}{DDPM} & MNIST         & 11.45    & 5.12         & 8.91       & 21.24      & 16.11       & 14.93           & \cellcolor[HTML]{BFBFBF}33.73     & 13.46   \\
                      & Fashion MNIST & 17.35    & 14.40        & 15.91      & 33.89      & 25.25       & 16.53           & \cellcolor[HTML]{BFBFBF}42.84     & 18.67   \\
                      & CIFAR-10      & 33.53    & 30.38        & 23.56      & 22.39      & 26.27       & 35.91           & \cellcolor[HTML]{BFBFBF}39.23     & 29.02   \\
                      & CIFAR-100     & \cellcolor[HTML]{BFBFBF}44.31    & 28.76        & 28.23      & 25.62      & 29.82       & 41.26           & \cellcolor[HTML]{BFBFBF}44.10     & 34.60   \\ \midrule
DDIM                  & CIFAR-100     & \cellcolor[HTML]{BFBFBF}70.16    & 52.33        & 67.35      & 61.23      & 56.88       & 68.55           & \cellcolor[HTML]{BFBFBF}71.63     & 65.60   \\ \bottomrule
\end{tabular}
}
\end{table}

\subsection{Severity Experiments}\label{severity_exp}
Our subsequent experiment sought to understand how the severity of fog corruption influenced the top-performing model, DDPM's return performance. We postulated that an escalation in corruption severity would result in a downturn in DDPM's performance. Experimental designs incorporated five tiers of fog severity $(\lambda=1 \sim 5)$ across five datasets (including the previously mentioned four and Tiny-ImageNet-C\cite{hendrycks2019benchmarking}). Using the scenario with $\lambda=1$ as a benchmark, we assessed the degree of performance degradation as severity fluctuated.

\begin{figure}[h]
  \centering
  \includegraphics[width=\textwidth]{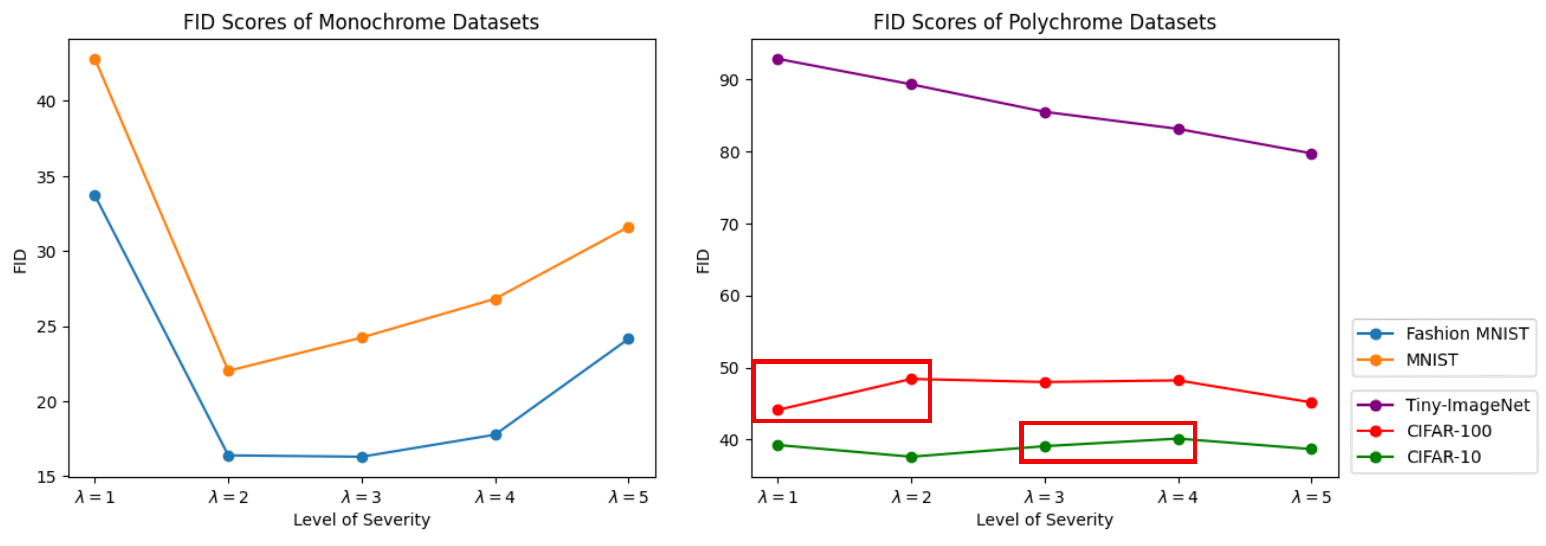}
  \caption{\textbf{The Variation in DDPM Performance with respect to Severity.} While fog corruption degrades the generative ability of DDPM, it doesn't get worse as it gets stronger.}
  \label{fig:figure6}
\end{figure}

In datasets with monochrome imagery, FID scores were surprisingly better (lower) at heightened corruption levels relative to the baseline, suggesting that fog corruption might have inadvertently aided DDPM in extracting features from monochrome images. 
For datasets with a broader color palette, no marked changes stood out. 
Minor deteriorations were observed at the initial shift in CIFAR-10 and the tertiary shift in CIFAR-100, though they were not of significant concern. 
While the intensity of fog severity appears to have a negligible effect on the Diffusion model's generative output, it was discerned that DDPM and DDIM's susceptibility to specific corruptions is notably shaped by topological and statistical congruences, especially the correspondence between mean and variance.

\subsection{Fractal Independence Experiments}\label{fractal_exp}
We delved deeper into the Fog corruption, rooted in the theoretical Plasma Fractal that showcases self-similarity traits and is generated via the Diamond-Square Algorithm\cite{fournier1982computer}. Given the fog's association with fractal designs, we hypothesized that DDPM and DDIM might encounter challenges when trained on images with fractal structures. To discern the notable generative constraints of DDPM and DDIM, we undertook an experiment employing the Fractal Dream dataset. This dataset is segmented into red, green, and blue image categories, each encompassing around 2,000 datasets at a resolution of $128 \times 128$ pixels. Before initiating training, we expanded this dataset using augmentation methods, amassing close to 22,000 samples.

\begin{table}[h!]
\centering
\caption{FID scores of DDPM on the Fractal Datasets}
\resizebox{\columnwidth}{!}{%
\begin{tabular}{@{}c|c|cccccccc@{}}
\toprule
     &       & Clear    & \multicolumn{2}{c}{Noise} & \multicolumn{2}{c}{Blur} & \multicolumn{2}{c}{Weather} & Extra   \\
     &       & Identity & Impulse      & Shot       & Glass      & Motion      & Brightness      & \cellcolor[HTML]{BFBFBF}Fog       & Spatter \\ \midrule
DDPM & \ Blue \  & 21.76    & 9.64         & 11.90      & 15.93      & 14.19       & 21.86           & \cellcolor[HTML]{BFBFBF}24.52     & 23.55   \\
     & \ Red \   & \cellcolor[HTML]{BFBFBF}26.16    & 14.96        & 9.86       & 13.98      & 22.63       & 23.96           & \cellcolor[HTML]{BFBFBF}26.19     & 18.67   \\
     & \ Green \ & 33.53    & 30.38        & 23.56      & 22.39      & 26.27       & 35.91           & \cellcolor[HTML]{BFBFBF}39.23     & 29.02   \\ \midrule
DDIM & \ Red \   & 95.73    & 69.05        & 65.56      & 85.54      & \cellcolor[HTML]{BFBFBF}129.46      & \cellcolor[HTML]{BFBFBF}98.74           & \cellcolor[HTML]{BFBFBF}99.66     & \cellcolor[HTML]{BFBFBF}120.71  \\
     & \ Green \ & 80.67    & 58.51        & 39.90      & 51.67      & 77.39       & 78.76           & \cellcolor[HTML]{BFBFBF}88.62     & \cellcolor[HTML]{BFBFBF}89.20   \\ \bottomrule
\end{tabular}
}
\end{table}

Surprisingly, our results indicated that neither DDPM nor DDIM demonstrated significant vulnerability in generating images with fractal patterns. We divided our analysis into two distinct cases: initially, we superimposed fractal corruption onto the foundational image $(\tilde{x}_0 = x_0 + \delta)$; subsequently, we considered images that intrinsically exhibited a fractal pattern $(x_{\delta})$. From this assessment, it became evident that, in particular, the generative prowess of DDPM might be considerably undermined when exclusively confronted with corruptions of a fractal nature.

\section{Discussion and Future works}
This research is focused on unravelling the inherent limitations of generative models. Employing the Diffusion-C framework, we evaluated the model's efficacy across an array of experimental conditions, leading us to three distinct conclusions (\uppercase\expandafter{\romannumeral1}, \uppercase\expandafter{\romannumeral2}, \uppercase\expandafter{\romannumeral3}). The trajectory of future research in this domain should address specific considerations: The enigmatic internal operations of deep learning often categorize it as a gray box, complicating our understanding of why generative models might be vulnerable to corruptions like the fractal one. In response, our approach has been to validate our propositions empirically by subjecting diffusion models to diverse environments. Furthermore, it's worth addressing the metric concern. Generative models aim to generate 'novelty'. 
However, as the FID predominantly measures similarity to existing data, there emerges a clear necessity for a new metric tailored to assess this element of novelty as well as corrupted ones.

\medskip
\newpage
\bibliographystyle{unsrt}
\bibliography{sample}

\newpage

\tableofcontents
\newpage
\appendix
\section{Introduction to Diffusion-C}\label{do_you_know_ddpm_c}
\subsection{Notations}

\begin{table}[ht!]
\caption{\textbf{Notations.}}
\captionsetup{justification=centering} 
\label{tab:notation}
\resizebox{0.9\columnwidth}{!}{
\begin{tabular}{@{}cl|cl@{}}
\toprule
\rowcolor[HTML]{EFEFEF} 
\textbf{Notations} & \multicolumn{1}{c|}{\cellcolor[HTML]{EFEFEF}\textbf{Description}} & \textbf{Notations} & \multicolumn{1}{c}{\cellcolor[HTML]{EFEFEF}\textbf{Description}} \\ \midrule
 $x$ & Non-corrupted observable data & {$\tilde{x}$} & Corrupted observable data\\ \midrule
{$x_0$} & Original point of $x$ &{$\tilde{x_0}$} & Original point of $\tilde{x}$  \\ \midrule
{$x_t$} & Diffused $x_0$ at time $t \in \{0, 1, \ldots, T\}$
 & {$\tilde{x_t}$} & Diffused $\tilde{x}_0$ at time $t \in \{0, \ldots, T\}$ \\ \midrule
{$x_T$} & Random noise after diffusing $x_0$ & {$\tilde{x_T}$} & Random noise after diffusing $\tilde{x_0}$\\ \midrule
{$x_g$} & Return point of $x_0$ &  {$\tilde{x_g}$} & Return point of $\tilde{x_0}$\\ \midrule
{$t$} & Time index $t \in \{0, 1, \ldots, T\}$
 &  {$y$} & target data\\ \midrule
{$D,G$} & Discriminator, Generator &  {$\theta , \phi$} & learnable parameters\\ \midrule
{$z$} & Random noise with normal distribution &  {$\mathcal{L}_D, \mathcal{L}_G$} & Loss function of $D$ and $G$\\ \midrule
{$D(G(z))$}& $P($\textit{real} $\vert$\textit{ fake image}$)$ & {$D(x)$} & $P($\textit{real} $\vert$ \textit{real image}$)$ \\ \midrule
{$L_0,L_{t-1}$} & Diffusion loss, denoising loss & {$L_T$} & Decoder loss \\ \midrule
{$\beta_t$} & Variance scale coeffficient & {$\alpha_t, \Bar{\alpha_t}$} &$\alpha_t := 1-\beta_t, \; \bar{\alpha_t} := \prod\limits_{s=1}^{t} \alpha_s$ \\ \midrule
{$\mathcal{C}$} & Corruptions set
 & {$\lambda$} & Severity Level of Corruptions.\\ \midrule
{$\epsilon$} & Gaussian noise
 & {$\delta_n$} & Each corruption for $\delta_k \in \mathcal{C}$\\ \midrule
{$\mathcal{N}$} & Normal distribution & {$\mathcal{F}, \tilde{\mathcal{F}}$} & Similarity between $(x_0,x_g)$ and $(\tilde{x_0},\tilde{x_g})$ \\ \midrule
{$q(x_t \vert x_{t-1})$} & Diffusion process at time $t$ & {$p_{\theta}(x_{t-1}\vert x_t)$} & Denoising process at time $t-1$ \\ \midrule
{$\epsilon_{\theta}(x_t,t)$} & Noise prediction model & {$\Sigma_{\theta}(x_t,t)$} & Variance coefficient of reversed step\\ 
\bottomrule
\end{tabular}%
}
\end{table}

\subsection{How does Diffusion-C work?}\label{ddpm_c_mechanism}

\begin{table}[h!]
\centering
\label{tab:my-table}
\resizebox{0.95\columnwidth}{!}{%
\begin{tabular}{@{}l@{}}
\toprule
\rowcolor[HTML]{EFEFEF} 
\textbf{Algorithm 1.} Mechanism on Diffusion-C \\ \midrule
\begin{tabular}[c]{@{}l@{}}1 : Input : $x_0$, $\;$Each corruptions : $\delta_k \tiny{(\delta_k \in \mathcal{C})}$, $\;$$\epsilon\sim\mathcal{N}(0,I)$, $\;$$\alpha_t := 1- \beta_t$ \end{tabular}\\
2 : \textbf{for} $\delta_k$ in $\mathcal{C}$ \textbf{do}                               \\
3 : $\qquad\tilde{x}_0 = x_t + \lambda \delta_k \ (1 \leq \lambda \leq 5)$  $\quad\quad\quad\quad\quad\quad\quad\quad\quad\quad\quad\quad\quad\quad\quad$\textit{// (1) corruption process}\\
4 : $\qquad\qquad$\textbf{repeat} \\
5 : $\qquad\qquad\tilde{x_0} \sim q(\tilde{x_0})$\\
6 : $\qquad\qquad$Take gradient descent step on \\
7 : $\qquad\qquad\qquad\nabla_{\theta}\vert\vert \epsilon - \epsilon_{\theta}(\sqrt{\Bar{\alpha_t}}\tilde{x_0} + \sqrt{1-\Bar{\alpha_t}}\epsilon, t)\vert\vert^2$\\
8 : $\qquad\qquad$\textbf{until} converged $\quad\quad\quad\quad\quad\quad\quad\quad\quad\quad\quad\quad$\textit{// (2) diffusion process}\\
9 : $\qquad\qquad$\textbf{for} $t = T,\ldots,1$ \textbf{do}\\
10 : $\qquad\qquad\qquad\tilde{x}_{t-1} = \frac{1}{\sqrt{\alpha_t}}(\tilde{x}_t - \frac{1-\alpha_t}{\sqrt{1-\Bar{\alpha_t}}}\epsilon_{\theta}(\tilde{x}_t,t)) + \sigma_t z$\\
11 :$\qquad\qquad$\textbf{end for} \\
12 :$\qquad\qquad\tilde{x}_g=\int p(\tilde{x}_T)\prod_{t=1}^{T}p_{\theta}(\tilde{x}_{t-1}\vert \tilde{x}_t) d\tilde{x}_{1:T} \quad\quad\,$ \textit{//   $\,$(3) denoising process}\\
13 :$\qquad$\textbf{return} $\max(\mathcal{F}(\tilde{x}_0, \tilde{x}_g),\mathcal{F}(x_0, x_g))\quad$\\
14 : \textbf{end for}\\
\bottomrule
\end{tabular}%
}
\end{table}

\begin{figure}[h!]
  \centering
  \includegraphics[width=0.9\textwidth]{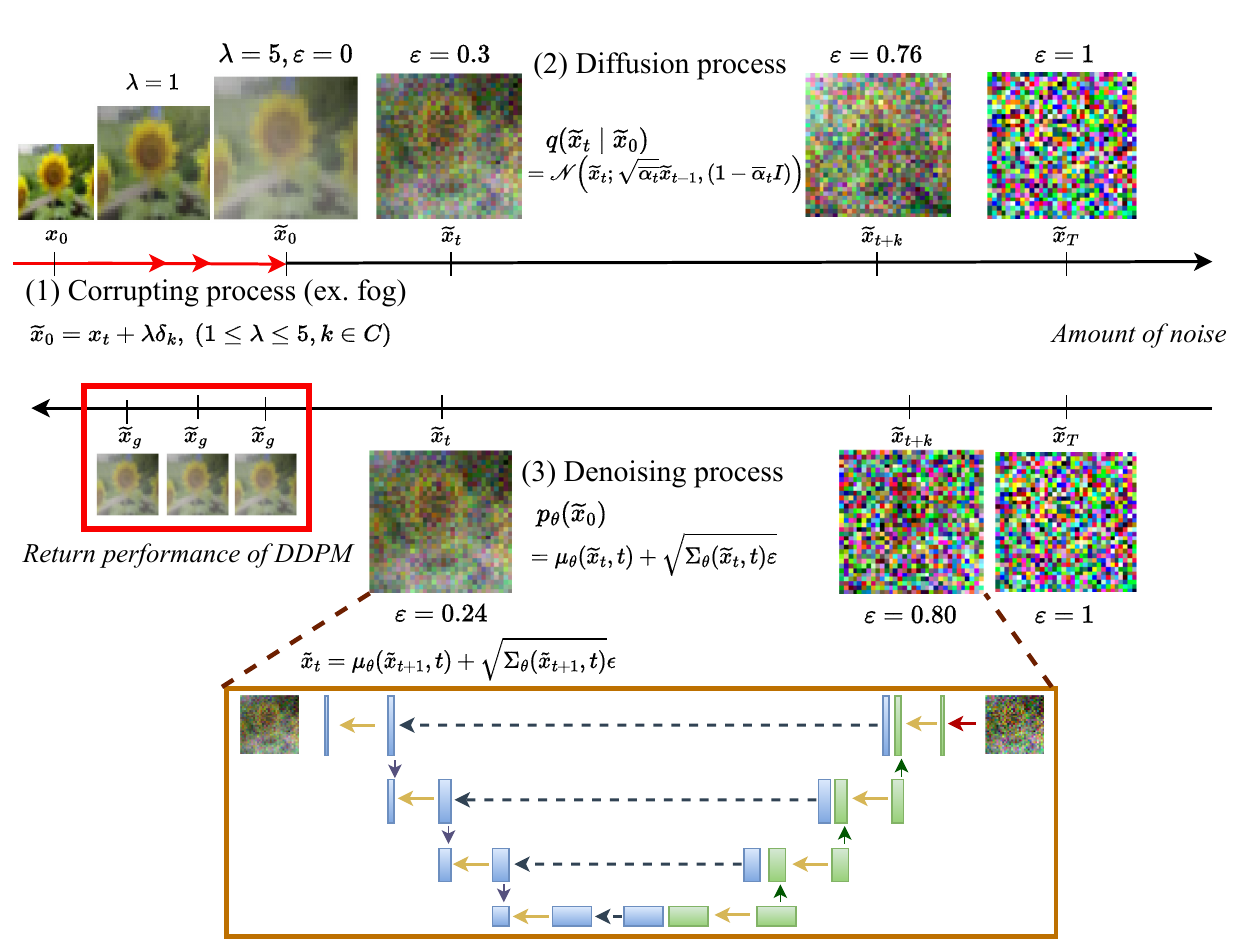}
  \caption{\textbf{Overall mechanism on Diffusion-C.} The training process of Diffusion-C consists of (1) Corrupting process, (2) Diffusion process and (3) Denoising process. For the denoising step, a U-Net\cite{ronneberger2015u} was used as the noise prediction network. After denoising process, Diffusion model returns final image $\tilde{x}_g$, which is the outcome of Diffusion-C. By calculating the differences between $\tilde{x}_0$ and $\tilde{x}_g$ in various conditional experiments, we identified when the return performance of DDPM deteriorates the most.}
  \label{fig:figure1}
\end{figure}

DDPM (Denoising Diffusion Probabilistic Models) \cite{ho2020denoising,cao2022survey,croitoru2023diffusion} is a generative model that employs the Markov Chain Monte Carlo (MCMC) method 
to learn the underlying structure of a dataset and generate a class of latent variable models. 
The objective of Diffusion-C is also to minimize the negative log-likelihood (NLL) function, which can be formulated as a minimization problem.

\begin{equation}
\begin{aligned}
E[-\log{p_{\theta}(\tilde{x}_0)}] &\leq E_q[-\log{\frac{p_{\theta}(\tilde{x}_{0:T})}{q(\tilde{x}_{1:T} \vert \tilde{x}_0)}}] = E_q[-\log{p(\tilde{x}_T)} - \sum_{t \geq 1} \frac{p_{\theta}(\tilde{x}_{t-1}\vert x_t)}{q(\tilde{x}_t \vert \tilde{x}_{t-1})}] \\
&= D_{KL}(q(\tilde{x}_T \vert \tilde{x}_0) \vert \vert p(\tilde{x}_T)) + \sum_{t > 1} D_{KL}(q(\tilde{x}_{t-1} \vert \tilde{x}_{t,0}) \vert \vert p_{\theta}(\tilde{x}_{t-1} \vert \tilde{x}_t, \tilde{x}_0))\\ 
&- E_q[\log{p_{\theta}(\tilde{x}_0 \vert \tilde{x}_1)}] = E_q[L_T + L_{t-1} + L_0] =: L.
\end{aligned}
\end{equation} 

Diffusion-C comprises of three stages: (1) Corrupting, (2) Diffusion, and (3) Denoising. 
Initially, the original dataset $x_0$ is corrupted into $\tilde{x}_0$, which results in the disturbance of the original point's information. 
Starting from the new original state, $\tilde{x}_0$ is transformed into $\tilde{x}_T$ through the diffusion process and eventually returns to $\tilde{x}_g$ through the denoising process. 
The effectiveness of DDPM is determined by the degree of similarity between $\tilde{x}_0$ and $\tilde{x}_g$, denoted by the expression $\mathcal{F}(\tilde{x}_0, \tilde{x}_g)$.

\textbf{(1) Corrupting Process($L_0$).} The corrupting process involves perturbing the original dataset with corruptions. 
Using various types and severities of corruptions, Diffusion-C transforms the original image into a corrupted one, $\tilde{x}_0$.
\begin{equation}
\tilde{x}_0 = x_t + \lambda \delta_k \ (0 \leq t \leq T, \ 1 \leq \lambda \leq 5, \ k \in \mathcal{C} ).
\end{equation}

\textbf{(2) Diffusion Process($L_T$)}. The diffusion process involves repeatedly adding Gaussian noise ($\beta_t$) to the corrupted image $\tilde{x}_t$ at every timestep $t$ to create a noisy state $\tilde{x}_T$. 
The diffusion process of Diffusion-C can be represented as $q(\tilde{x}_{1:T} \vert \tilde{x}_0 )$. 
\begin{equation}
q(\tilde{x}_t \vert \tilde{x}_0) = \mathcal{N}(\tilde{x}_t ; \sqrt{\Bar{\alpha_t}} \tilde{x}_0, (1-\Bar{\alpha_t})I).
\end{equation}

\textbf{(3) Denoising Process ($L_{1:T-1}$)}. DDPM trains the denoising process to move $\tilde{x}_T$ back to the original state of $\tilde{x}_0$. 
In this process, the noise injected into the image is removed to reveal the original state. 
It is represented by $ p_{\theta}(\tilde{x}_{0:T}) := p(\tilde{x}_T)\prod_{t=1}^{T} p_{\theta}(\tilde{x}_{t-1} \vert \tilde{x}_t)$, and the denoising process at time $t$ can be expressed as follows.

\begin{equation}
    \begin{aligned}
       p_{\theta}(\tilde{x}_{t-1} \vert \tilde{x}_t):= N(\tilde{x}_{t-1}; \mu_{\theta} (\tilde{x}_t),\Sigma_{\theta} (\tilde{x}_t,t)). \\
    \end{aligned}
\end{equation}

$p_{\theta}$ is used to learn the process of removing the noise based on $q$ which learns $\mu_{\theta}(\tilde{x}_t,t)$ and $\Sigma_{\theta}(\tilde{x}_t,t)$.

\section{Detailed Experimental Explanations}
\subsection{Basic Comparison Experiment}\label{detailed_comparison_exp}

\textbf{Type of Corruptions.} In this experiment, we used total seven corruptions. Shot noise is the Poisson noise, and Impulse noise is the salt-pepper noise. Glass blur simulates noises through glass by mixing pixels. Motion blur is the corruption blurring the image along any line. Brightness is made with the increment of images. Fog simulates fog using diamond-square algorithm. Spatter is a corruption that simulates when an image is wet with water.
\begin{figure}[h]
  \centering
  \includegraphics[width=0.8\textwidth]{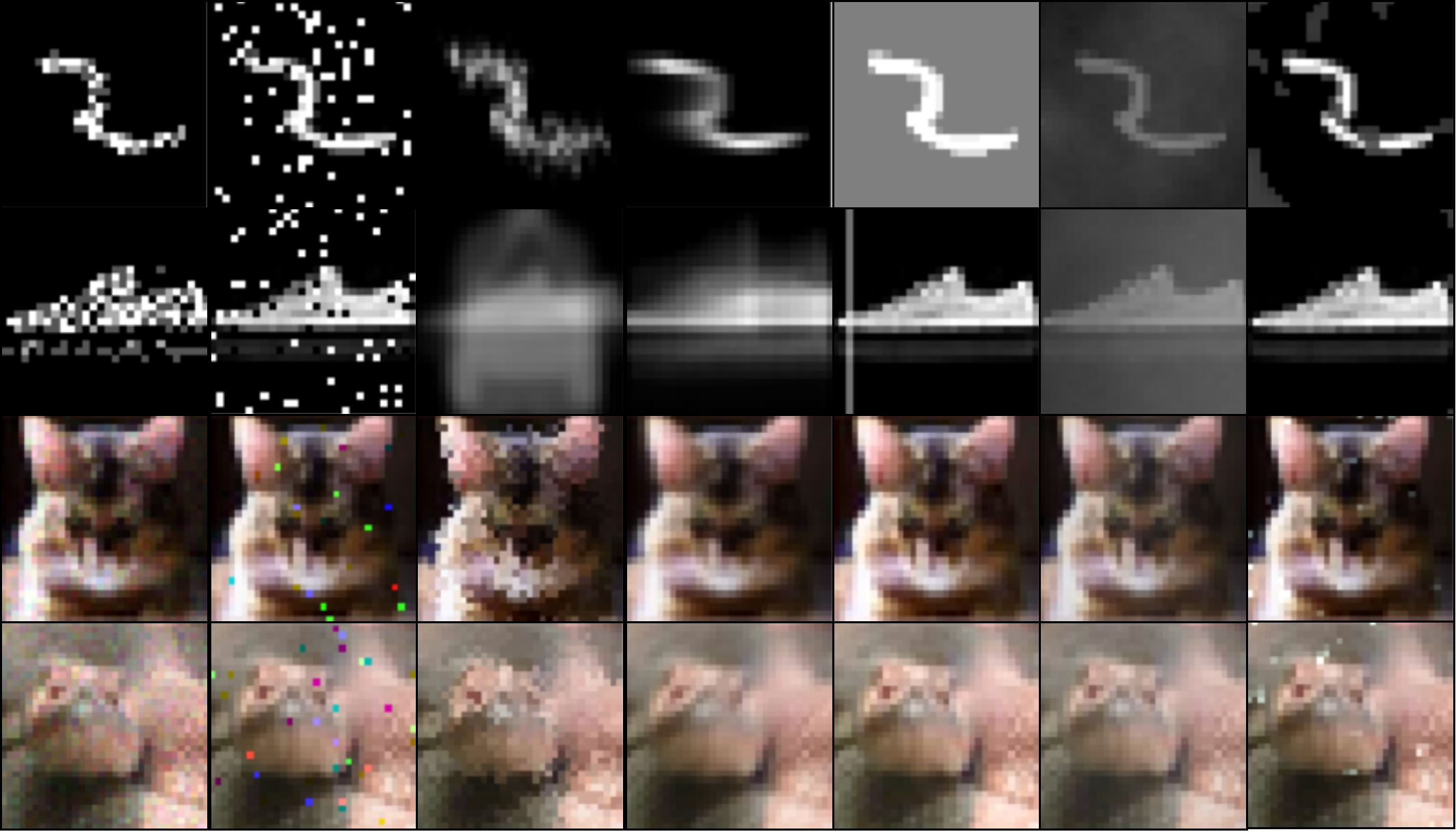}
  \caption{\textbf{Seven corruptions.} In eight corruptions, there are shot noise, impulse noise, glass blur, motion blur, brightness, fog, and spatter.}
  \label{fig:figure6}
\end{figure}

\textbf{Results of DCGAN\cite{radford2015unsupervised}.} Among the three models evaluated, DCGAN performed the worst. 
While it achieved a relatively small FID score on the identity dataset, its performance degraded gradually as the dataset became corrupted.
DCGAN experienced mode collapse in the presence of impulse noise or shot noise, resulting in repetitive features in generated images. 
This suggests that the model struggled with the challenges posed by impulse noise and shot noise during training.

\begin{table}[h]
\centering
\caption{Results of DCGAN}
\label{tab:my-table}
\resizebox{\columnwidth}{!}{%
\begin{tabular}{lcccccccc}
\hline
\rowcolor[HTML]{EFEFEF} 
\multicolumn{1}{c}{\cellcolor[HTML]{EFEFEF}} &
  Clear &
  \multicolumn{2}{c}{\cellcolor[HTML]{EFEFEF}Noise} &
  \multicolumn{2}{c}{\cellcolor[HTML]{EFEFEF}Blur} &
  \multicolumn{2}{c}{\cellcolor[HTML]{EFEFEF}Weather} &
  Extra \\ \cline{2-9} 
\rowcolor[HTML]{EFEFEF} 
\multicolumn{1}{c}{\multirow{-2}{*}{\cellcolor[HTML]{EFEFEF}}} &
  identity &
  impulse &
  shot &
  glass &
  motion &
  fog &
  brightness &
  spatter \\ \hline
MNIST    & 61.48 & \cellcolor[HTML]{C0C0C0}251.80 & 64.25 & 72.65 & 105.53 & 61.54 & 68.82 & 101.99 \\
FashionMNIST & 46.14 & \cellcolor[HTML]{C0C0C0}249.99 & 53.02 & 125.92 & 191.63 & 94.69 & 51.69 & 59.35 \\
CIFAR-10  & 189.13 & \cellcolor[HTML]{C0C0C0}304.68 & 190.53 & 209.24 & 151.07 & 189.93 & 231.99 & 238.97 \\
CIFAR-100 & 190.74 & \cellcolor[HTML]{C0C0C0}228.49 & 164.11 & 190.65 & 195.47 & \cellcolor[HTML]{C0C0C0}257.49 & 211.35 & \cellcolor[HTML]{C0C0C0}244.30 \\ \hline
\end{tabular}%
}
\end{table}

\textbf{Results of WGAN-GP\cite{gulrajani2017improved}.} Unlike DCGAN, WGAN-GP generated images with diverse features and reduced the performance gap between monochrome and polychrome datasets. 
However, it still faced challenges in feature extraction and generation when working with corrupted datasets.

\begin{table}[h!]
\centering
\caption{Results of WGAN-GP}
\label{tab:my-table}
\resizebox{\columnwidth}{!}{%
\begin{tabular}{lcccccccc}
\hline
\rowcolor[HTML]{EFEFEF} 
\multicolumn{1}{c}{\cellcolor[HTML]{EFEFEF}} &
  Clear &
  \multicolumn{2}{c}{\cellcolor[HTML]{EFEFEF}Noise} &
  \multicolumn{2}{c}{\cellcolor[HTML]{EFEFEF}Blur} &
  \multicolumn{2}{c}{\cellcolor[HTML]{EFEFEF}Weather} &
  Extra \\ \cline{2-9} 
\rowcolor[HTML]{EFEFEF} 
\multicolumn{1}{c}{\multirow{-2}{*}{\cellcolor[HTML]{EFEFEF}}} &
  identity &
  impulse &
  shot &
  glass &
  motion &
  fog &
  brightness &
  spatter \\ \hline
MNIST    & 96.40 & \cellcolor[HTML]{C0C0C0}133.60 & 95.99 & 86.61 & 108.52 & \cellcolor[HTML]{C0C0C0}140.51 & 95.51 & \cellcolor[HTML]{C0C0C0}150.04 \\
FashionMNIST & 110.02 & \cellcolor[HTML]{C0C0C0}136.92 & 70.64 & 68.72 & 128.79 & \cellcolor[HTML]{C0C0C0}144.12 & 97.08 & 99.07 \\
CIFAR-10  & 126.25 & 113.48 & 117.75 & 113.98 & \cellcolor[HTML]{C0C0C0}136.57 & 100.62 & 119.96 & \cellcolor[HTML]{C0C0C0}171.68 \\
CIFAR-100 & 134.46 & 120.93 & \cellcolor[HTML]{C0C0C0}153.76 & 132.79 & 98.73 & 94.50 & 123.13 & \cellcolor[HTML]{C0C0C0}152.93 \\ \hline
\end{tabular}%
}
\end{table}

\newpage
\subsection{Severity Experiment}\label{detailed_severity_exp}

In this experiment, we mainly used and analyzed the impact of fog corruptions.
\begin{figure}[h!]
  \centering
  \includegraphics[width=\textwidth]{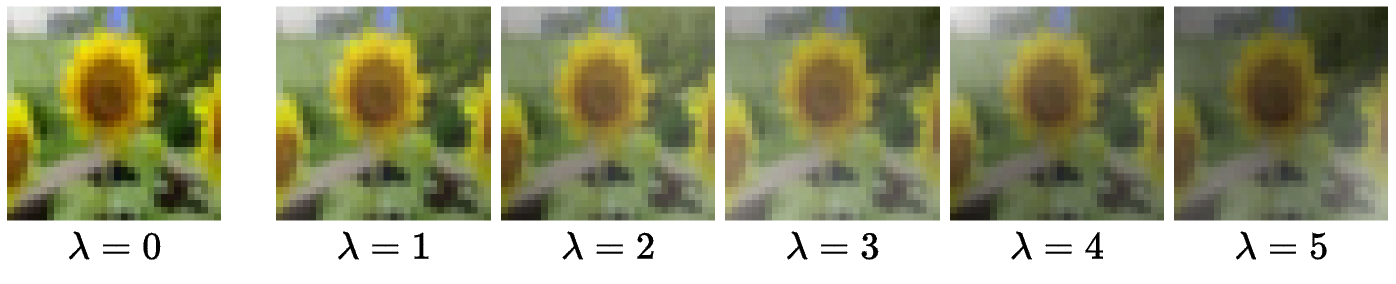}
  \caption{\textbf{Image changes with severity.} In this paper, we observed how the performance of DDPM degrades as the severity of the fog corruption increases by varying the severity from 1 to 5.}
\end{figure}

\begin{figure}[h!]
  \centering
  \includegraphics[width=\textwidth]{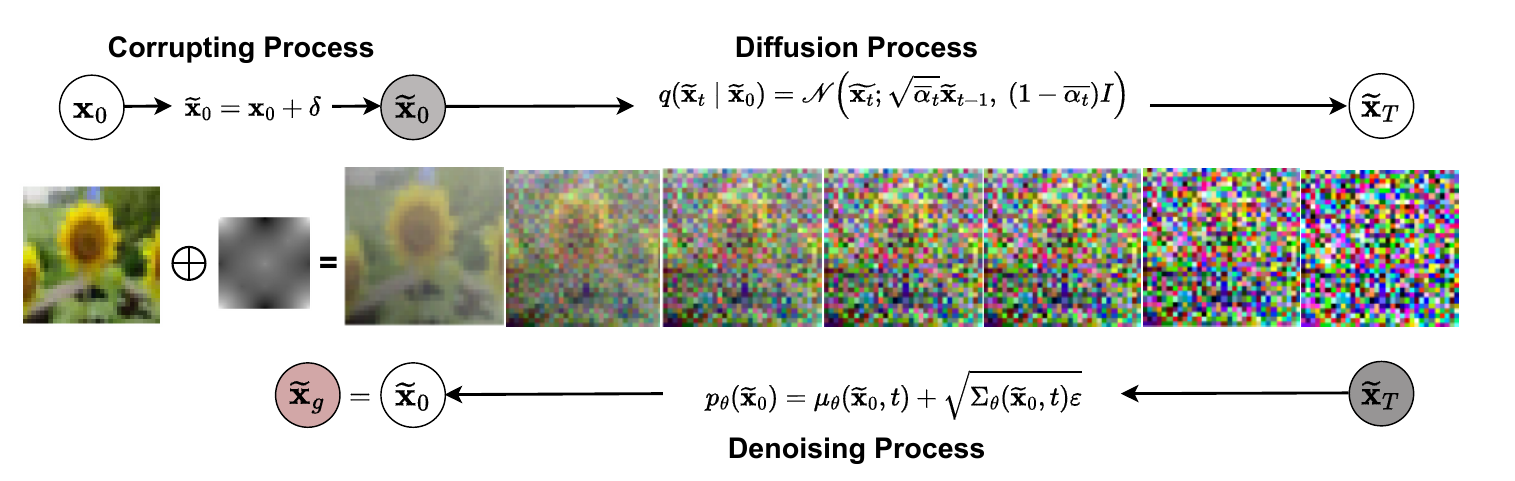}
  \caption{\textbf{Mechanism of Diffusion-C used in the Severity Experiments.}}
  \label{fig:figure1}
  \label{fig:figure1}
\end{figure}

\begin{table}[h!]
\centering
\caption{FID Scores based on the variations of severity ($\lambda$).}
\label{tab:my-table}
\resizebox{0.7\columnwidth}{!}{%
\begin{tabular}{lcccccccc}
\hline
\rowcolor[HTML]{EFEFEF} 
\multicolumn{1}{c}{\cellcolor[HTML]{EFEFEF}}
              & $\lambda=1$     & $\lambda=2$     & $\lambda=3$     & $\lambda=4$     & $\lambda=5$     \\ \hline
MNIST         & 33.70 & 16.39 & 16.30 & 17.77 & 24.15 \\
Fashion MNIST & 42.80 & 22.02 & 24.24 & 26.82 & 31.61 \\
CIFAR-10      & 39.23 & 37.60 & 39.07 & 40.12 & 38.66 \\
CIFAR-100     & 44.11 & 48.41 & 47.98 & 48.21 & 45.15 \\
Tiny-ImageNet & 92.88 & 89.34 & 85.52 & 83.15 & 79.15 \\ \bottomrule
\end{tabular}
}
\end{table}

\subsection{Fractal Independence Experiment}\label{detailed_fractal_exp}
In this experiment, we compared the performance of DDPM between the input as the images added independent fractal corruptions and original fractal images.
\begin{figure}[h!]
  \centering
  \includegraphics[width=\textwidth]{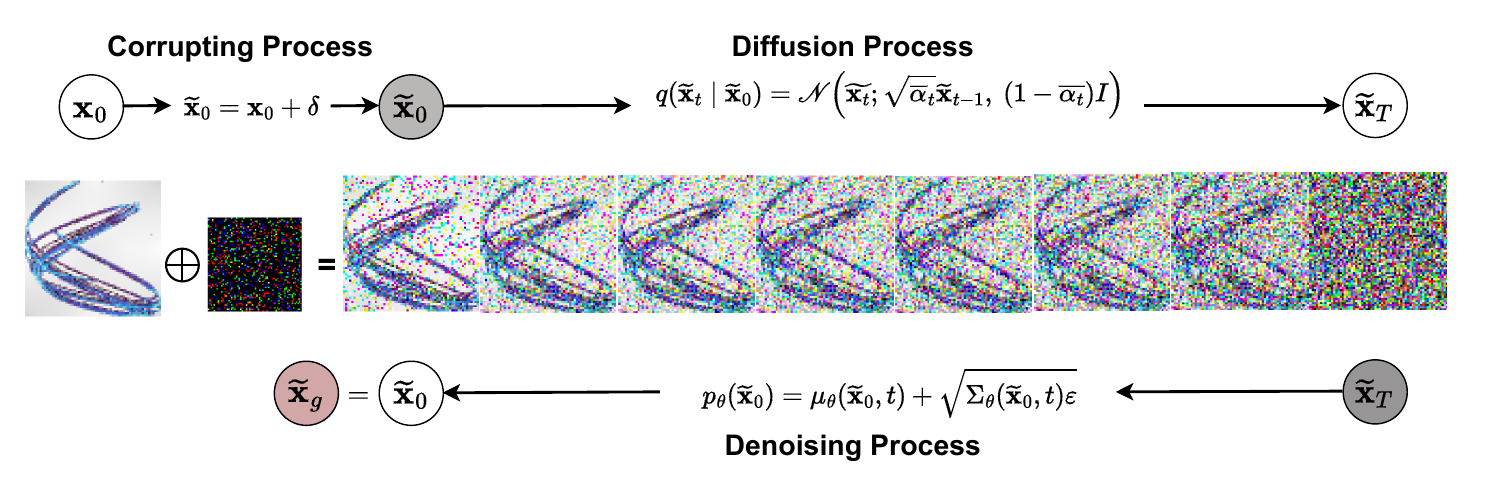}
  \caption{\textbf{Mechanism of Diffusion-C used in Fractal Independence experiment.} }
  \label{fig:figure1}
\end{figure}

\end{document}